\documentclass[letterpaper, 10 pt, conference]{ieeeconf}  % Comment this line out if you need a4paper

\IEEEoverridecommandlockouts                              % This command is only needed if 
\usepackage{cite}
\usepackage{amsmath,amssymb,amsfonts}
\usepackage{algorithmic}
\usepackage{graphicx}
\usepackage{textcomp}
\usepackage{gensymb}
\usepackage{xcolor}
\usepackage{color}		
\usepackage{multirow}		
\usepackage{array}										
\usepackage{colortbl}
\usepackage{siunitx}
\usepackage[english]{babel}	
\usepackage{multirow}
\usepackage[hidelinks]{hyperref}     
\usepackage{blindtext}
\usepackage[ruled, linesnumbered]{algorithm2e}

\title{\LARGE \bf
Intuitive Telemanipulation of Hyper-Redundant Snake Robots within \mbox{Locomotion and Reorientation using Task-Priority Inverse Kinematics}
}

\author{Tim-Lukas Habich, Melvin Hueter, Moritz Schappler and Svenja Spindeldreier$^{1}$%
\thanks{$^{1}$All authors are with the Leibniz University Hannover, Institute of Mechatronic Systems, \mbox{30823 Garbsen, Germany,}
{\tt tim-lukas.habich@imes.uni-hannover.de}
}
}

\newif\ifcopyright
	\copyrighttrue

\newcommand{\mm}[1]{\boldsymbol{#1}}
\newcommand{\m}[5]{{}_{\mathrm{#2}}^{\mathrm{#3}}{\mm{#1}}^{\mathrm{#4}}_{\mathrm{#5}}}

\newcommand{\e}[2]{\begin{equation} #1 \label {eq:#2} \end{equation}}

\newcommand{\ind}[1]{\mathrm{#1}}
\usepackage[mathscr]{euscript}
\newcommand{\fra}[1]{{\mathscr{F}}_{#1}}
\newcommand{\tJac}[1]{\mm J_{\mathcal{T}#1}}
\definecolor{Gray}{gray}{0.85}

\newcolumntype{M}[1]{>{\centering\arraybackslash}m{#1}}
\newcolumntype{N}{@{}m{0pt}@{}}
\makeatletter
\newcommand{\removelatexerror}{\let\@latex@error\@gobble}
\makeatother

\begin{document}
\ifcopyright
{\LARGE IEEE Copyright Notice}
\newline
\fboxrule=0.4pt \fboxsep=3pt

\fbox{\begin{minipage}{1.1\linewidth}  % <-- hier Kastenbreiter der Kopfzeile ändern
		%\centering
		% <-- hier Namen der Konferenz und Jahr einfügen
		%Changes were made to this version by the publisher prior to publication. \\
		%	The final version of record is available at http://doi.org/10.1109/XYZ123456789.TODO   % <-- hier DOI einfügen
		\textcopyright\,\,2023\,\,IEEE. Personal use of this material is permitted. Permission from IEEE must be obtained for all other uses, in any current or future media, including reprinting/republishing this material for advertising or promotional purposes, creating new collective works, for resale or redistribution to servers or lists, or reuse of any copyrighted component of this work in other works. \\
		
		Accepted to be published in: Proceedings of the 2023 IEEE International Conference on Robotics and Automation (ICRA), May 29 -- June 2, 2023, London, England.\\
		
		DOI: 10.1109/ICRA48891.2023.10161124
		
\end{minipage}}
\else
\fi
\graphicspath{{./images/}}

\maketitle
\thispagestyle{empty}
\pagestyle{empty}

%%%%%%%%%%%%%%%%%%%%%%%%%%%%%%%%%%%%%%%%%%%%%%%%%%%%%%%%%%%%%%%%%%%%%%%%%%%%%%%%
\begin{abstract}
Snake robots offer considerable potential for endoscopic interventions due to their ability to follow curvilinear paths. Telemanipulation is an open problem due to hyper-redundancy, as input devices only allow a specification of six degrees of freedom. Our work addresses this by presenting a unified telemanipulation strategy which enables follow-the-leader locomotion \textit{and} reorientation keeping the shape change as small as possible. The basis for this is a novel shape-fitting approach for solving the inverse kinematics in only a few milliseconds. Shape fitting is performed by maximizing the similarity of two curves using Fréchet distance while simultaneously specifying the position and orientation of the end effector. Telemanipulation performance is investigated in a study in which 14 participants controlled a simulated snake robot to locomote into the target area. In a final validation, pivot reorientation within the target area is addressed.
\end{abstract}
\section{Introduction}
\begin{figure}[tbp]
		\centering
		\resizebox{1\linewidth}{!}{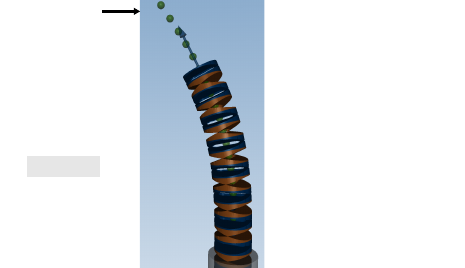}
		\caption{(a) Intuitive telemanipulation with SnakeTTP using an input device (frames: base $\fra{\mathrm{B}}$, stylus $\fra{\mathrm{S}}$). The snake's tip is controlled by orientation input $\m{R}{}{B}{}{S}$. A camera at the distal end provides visual feedback. Buttons activate the feeder for locomotion ($b_1$) or realize pivot reorientation ($b_2$). Green spheres indicate a desired curvilinear path to be followed. (b) Less shape change is realized during reconfiguration compared to a reference.} \label{fig:cover}
\end{figure}

Minimally invasive surgery (MIS) has changed medicine sustainably. Intraluminal procedures offer a reduction of incisions and postoperative pain which accelerates recovery. This is made possible by access via natural orifices, resulting in two key demands: Endoscopes require maneuverability to reach the target area and sufficient stiffness for tissue manipulation. Conventional flexible endoscopes have high compliance at the actuated distal end due to the shaft's passive structure \cite{Vitiello.2013}. Identical problems occur in the boroscopy, e.g. of aircraft engines. Reaching the compressor blades is possible due to the flexible tube, but the necessary stiffness for surface processing is lacking. One solution is a change from a passive shaft to a fully actuated structure. Continuum robots \cite{BurgnerKahrs.2015} and hyper-redundant (discrete) snake robots \cite{Chirikjian.1992}, as the one shown in Fig.~\ref{fig:cover}(a), fulfill these requirements. Numerous works focus on snake-motion imitation, e.g. realizing sidewinding and lateral undulation via environmental contact \cite{Liljeback.2005} or the repulsion at the environment \cite{Transeth.2008}. These approaches are not applicable for MIS: contact with the tissue should be reduced to avoid injuries. Locomotion should be provided by an external feeder and minimization of tissue contact realized by \emph{fitting the snake's shape} to the anatomical pathway.

For shape control, snake robots can be modeled using a continuous backbone curve that takes into account macroscopic geometric features \cite{Chirikjian.1992,Yamada.2006}. In \cite{Mochiyama.1999}, a parameterized spatial curve is used to specify a desired shape. A similar approach generates the desired shape by interpolating shape control points and successively aligning the robot along the curve \cite{Liljeback.2014}. Tang et al. \cite{Tang.2020} also use waypoints as input, form the target shape using spline interpolation and determine the joint angles with kinematic considerations. Alternatively, an optimization problem can be formulated to minimize the distances between the current link positions and target positions on a given backbone \cite{Wang.2021}. All these approaches are not suitable for telemanipulating snake robots for one reason: the assumption of a known desired snake shape. The target shape is not known a priori and must be created incrementally via user input. This is true both during locomotion into the target area and during \emph{reorientation within the target area}, as pivot movements will inherently change the shape of the robot.

To counter this, follow-the-leader (FTL) motion was formulated \cite{WilliamsII.1997}. Controlling the snake tip by a joystick enables whole-body collision avoidance by having all links spatially follow the tip. Many groups \cite{Tappe.2015,Choset.1999,Xie.2019} have researched FTL movements for MIS or inspection tasks based on this idea. Palmer et al. \cite{Palmer.2014} propose a tip-following method for continuum robots using sequential quadratic programming. The user can control the tip via a two-axis gamepad and path-tracking is implemented using a geometric cost function.

However, telemanipulating via gamepads or joysticks with two degrees of freedom (DoF) in a classical FTL fashion \cite{WilliamsII.1997} is not promising for two reasons: First, such devices allow only rudimentary haptic feedback (vibration). A precise feedback of the 3D interaction force between instrument and tissue is impossible. This is essential for robot-assisted surgery, as it reduces intraoperative injuries and improves surgical performance \cite{H.XinJ.S.ZelekandH.Carnahan.2006}. Kwok et al. \cite{Kwok.2013} therefore use a six-DoF haptic input device for telemanipulation and realize motion planning with dynamic active constraints. To consider tissue deformation, pre- and intraoperative imaging (e.g. computer tomography of the entire environment) is required. Such imaging is not always available and telemanipulation of snake robots for MIS or inspection of industrial machines should be possible without such external sensing.

%%%%%%%%%%%%%%%%VIDEO
%\begin{figure}[t]
%	\removelatexerror
%	\vspace{2mm}
%	\begin{algorithm}[H]
%		\caption{SnakeTTP}\label{alg:snakettp}
%		{\small 
%			\SetKwInOut{Input}{Input}
%			\SetKwInOut{Output}{Output}
%			\Input{$\mm{q},b_1,b_2,\m{R}{}{B}{}{S}$}
%			\Output{${\mm q}_{\mathrm{d}}$}
%			$\mm{\phi}_\mathrm{xy}\gets$ Pitch and yaw angles of $\m{R}{}{B}{}{S}$\;
%			$\mm{\kappa}_\mathrm{state}\gets$ Set states of active joints to 1\;
%			\uIf{$b_1$}
%			{\tcp{FTL locomotion}
%				$\mm{q}\gets$ Update pointing direction using $\mm{\phi}_\mathrm{xy}$\;
%				$\mm{P}_{\mathrm{d}}\gets$ Compute desired link positions \cite{WilliamsII.1997}\;
%				$^{0}\mm{T}_\ind{E,d}\gets$ Compute desired EE pose with $\mm{\phi}_\mathrm{xy}$ and $\mm{P}_{\mathrm{d}}$\;
%				${\mm q}_{\mathrm{d}}\gets$ Shape fitting (point-to-point correspondences)\;
%			}
%			\uElseIf{$b_2$}
%			{\tcp{Pivot reorientation}
%				$\mm{P}_{\mathrm{d}}\gets$ Compute current link positions with $\mm q$\;
%				$^{0}\mm{T}_\ind{E,d}\gets$ Compute desired EE pose with $\mm{\phi}_\mathrm{xy}$ and $\mm{P}_{\mathrm{d}}$\;
%				${\mm q}_{\mathrm{d}}\gets$ Shape fitting (Fréchet distance)\;
%			}
%			\Else{
%				\tcp{Telemanipulate last module}
%				$\mm{q}\gets$ Update pointing direction using $\mm{\phi}_\mathrm{xy}$\;
%				${\mm q}_{\mathrm{d}}\gets\mm{q}$\;
%			}
%		}
%	\end{algorithm}
%	\vspace{-6mm}
%\end{figure}
%%%%%%%%%%%%%%%%VIDEO

Second, a reorientation of the snake tip at a \emph{constant} end-effector (EE) position is not possible during FTL. Such a pivot motion (with instrument tip as pivot point) is important if the target area (e.g. a polyp to be ablated) is outside the camera view and the \emph{pointing direction} should be adjusted. This reconfiguration task requires online-planned shape fitting with consideration of position and orientation constraints. In \cite{BerthetRayne.2018}, six different approaches for telemanipulating snake robots during ear-nose-throat surgery are compared and promising results regarding real-time capability and intuitiveness are achieved using the pseudo-inverse of the Jacobian. Due to mechanical coupling, there is only a low degree of redundancy (robot with eight actuated DoF). The tip is controlled by a six-DoF interface, though there is no formulation of shape adaptation for whole-body collision avoidance. The pivot movement is also not addressed.

%Kwok et al. \cite{Kwok.2013} use the shape control approach of \cite{Mochiyama.1999} and a haptic input device for telemanipulation. Motion planning is performed using dynamic active constraints, such that a safety manipulation margin exists for the entire robot structure using a real-time modeled volumetric path. This changes according to tissue deformation and thus pre- and intraoperative imaging (e.g. computer tomography of the entire tissue environment) is required. However, such imaging is not always available and telemanipulation of snake robots for MIS or inspection of industrial machines should be able to be performed without such external sensing.

Intuitive telemanipulation of highly articulated structures is still considered an open problem in robotics \cite{Orekhov.2018} and the core of our work, illustrated in Fig.~\ref{fig:cover}. Our contributions are 
\begin{enumerate}
	\item formulating a novel shape-fitting method using task-priority inverse kinematics \cite{Baerlocher.1998},
	\item defining two different shape-fitting tasks using point-to-point correspondences and Fréchet distance and
	\item presenting a unified method for telemanipulating snake robots for locomotion \textit{and} reorientation (Algorithm~\ref{alg:snakettp}).
\end{enumerate}
Using the Fréchet distance in motion planning of redundant robots is not entirely new. Within \cite{Holladay.2016,Holladay.2019}, path planning is performed such that a desired task-space path is followed using the Fréchet distance, even though obstacles are present. However, the authors only consider the EE path of a seven-DoF manipulator which is not sufficient for hyper-redundant snake robots. The latter requires setting a six-DoF EE pose while simultaneously maintaining a desired snake shape to realize whole-body collision avoidance. This is the core of our work. To the best of the authors' knowledge, this is the first time that shape fitting is solved using the Fréchet distance as similarity measure between the desired and the current backbone. The paper is structured as follows: Section~\ref{FTL-tele} describes our telemanipulation strategy, followed by validations (Sec.~\ref{val}) and conclusions (Sec.~\ref{conclusions}).
\section{Snake-Robot Telemanipulation}\label{FTL-tele}
After formulating the kinematics (\ref{forwardkine}), shape fitting with orientation constraints (\ref{shapefitting}) and the online motion planning (\ref{ftlform}) are presented using a snake robot with alternating single-axis pitch and yaw joints. The method is easily extensible to robots with universal (two-DoF) joints.
\subsection{Kinematics Preliminaries}\label{forwardkine}
\begin{figure}[t]
	\centering
	\vspace{3mm}
	\resizebox{\linewidth}{!}{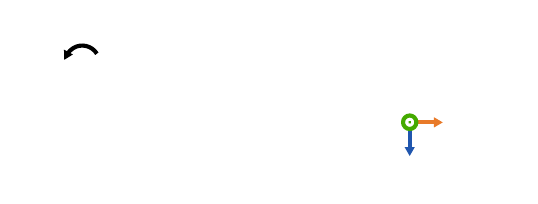}
	\caption{Kinematic chain with $n$ rotational actuators of height $h$, linear feeder $q_1$, frames $\fra{i}$ and arbitrary tool transformation $^{n{+}1}\mm{T}_\ind{E}$} \label{fig:kinematic_chain}
\end{figure}
\renewcommand{\arraystretch}{1.2}
\begin{table}[b]
	%\vspace{1.5mm}
	\caption{DH parameters of the snake robot}
	\vspace{-4mm}
	\label{tab:DH_param}
	\begin{center}
		\begin{tabular}{|c|c|c|c|c|}
			%\rule{0pt}{0ex}
			\hline
			$i$&$\theta_i$&$d_i$&$a_i$&$\alpha_i$\\
			\hline
			$1$&$0$&$q_1+h/2$&$0$&$-\pi/2$\\
			\hline
			$2$&$q_2-\pi/2$&$0$&$h$&$\pi/2$\\
			\hline
			$3$&$q_3$&$0$&$h$&$-\pi/2$\\
			\hline
			$4$&$q_4$&$0$&$h$&$\pi/2$\\
			\hline
			\vdots&\vdots&\vdots&\vdots&\vdots\\
			\hline
			$n-1$&$q_{n-1}$&$0$&$h$&$-\pi/2$\\
			\hline
			$n$&$q_{n}$&$0$&$h$&$\pi/2$\\
			\hline
			$n+1$&$q_{n+1}$&$0$&$h/2$&$0$\\
			\hline		
		\end{tabular}
	\end{center}
	\vspace{-1.5mm}	
\end{table}
Snake robots with $n$ discrete revolute joints are a serial chain of actuators. The Denavit-Hartenberg (DH) notation can be used to determine forward kinematics. Figure~\ref{fig:kinematic_chain} illustrates the structure neglecting the actuation principle. A feeder $q_\mathrm{1}$ for locomotion leads to a robot with \mbox{$(n{+}1)$-DoF}. The longitudinal rotation $\zeta{=}\pi/2$ between each joint allows spatial motion for each module consisting of two actuators.
The DH parameters for $n/2$ modules are given in Table~\ref{tab:DH_param}. The transformation $^{0}\mm{T}_i(\mm{q}_{i})$ depends on the first~$i$ coordinates $\mm{q}_i{=}[q_1,\hdots,q_i]^\ind{T}$ and contains the orientation $^{0}\mm{R}_i$ and position $\mm{p}_{i}$ of link $i$.
%\e{ ^{i-1} \mm{T}_i(q_i)= \left[ \begin{array}{cccc}
%		\ind{c}_{\theta_i} & -\ind{s}_{\theta_i} \, \ind{c}_{\alpha_i} & \ind{s}_{\theta_i} \, \ind{s}_{\alpha_i} & a_i \, \ind{c}_{\theta_i} \\
%		\ind{s}_{\theta_i} & \ind{c}_{\theta_i} \, \ind{c}_{\alpha_i} & -\ind{c}_{\theta_i} \, \ind{s}_{\alpha_i} & a_i \, \ind{s}_{\theta_i} \\
%		0 & \ind{s}_{\alpha_i} & \ind{c}_{\alpha_i} & d_i \\
%		0 & 0 & 0 & 1 \end{array}\right]}{Ti}
%applies ($\ind{c}_{\diamond_i}=\cos(\diamond_i)$ and $\ind{s}_{\diamond_i}=\sin(\diamond_i)$). By product formation we obtain the transformation
%\e{^{0}\mm{T}_i(\mm{q}_{i})=\prod_{j=1}^{i} {^{j-1}\mm{T}_{j}}(q_{j}) =\left[ \begin{array}{cc}
%		^{0}\mm{R}_i & \mm{r}_i \\
%		\mm{0} & 1 \end{array}\right]}{forward_kine}
%depending on the first $i$ joint positions $\mm{q}_i=[q_1,\hdots,q_i]^\ind{T}$.
The six-DoF pose is denoted as
\e{\mm x_i=\mm f_i(\mm{q}_i)=\left[ \begin{array}{c}
		\mm{p}_{i}\\
		\mm{\varphi}_\mathrm{xyz}(^{0}\mm{R}_i)\end{array}\right]\in \mathbb{R}^{6}.}{fk}
Without loss of generality, the (intrinsic) x-y'-z'' Tait-Bryan angles $\mm{\varphi}_\mathrm{xyz}$ represent orientation. 
%Consequently, by time differentiation the link velocity results in
%\e{\dot{\mm x}_i= \frac{\partial\mm x_i}{\partial\mm{q}_i}\dot{\mm{q}}_i=\mm J_i\dot{\mm{q}}_i}{diffk}
%with Jacobian $\mm J_i\in \mathbb{R}^{m_i\times i}$ of link $i$. 
For positioning link $i$, the inverse kinematics solution is first defined on velocity level. We distinguish between the analytical Jacobian~$\mm J_i{=}[\mm{J}_{i,\ind{t}}^\mathrm{T},\mm{J}_{i,\ind{r}}^\mathrm{T}]^\ind{T}\in\mathbb{R}^{6\times i}$ (\mbox{$\ind{t}$: translational,} \mbox{$\ind{r}$: rotational}) and the Jacobian~\mbox{$\tilde{\mm{J}}_i\in\mathbb{R}^{6\times i}$} of the task residual $\tilde{\mm x}_i$. Since for a 6D pose specification the na\"ive difference between the feedback task coordinate $\mm x_i$ and the desired quantity $\mm x_{i,\mathrm{d}}$ is only appropriate for position and not for orientation errors, the two different Jacobians are necessary \cite{Natale.2003}. By Moore-Penrose pseudo-inversion ($\dagger$), the desired minimum-norm joint angle velocities are obtained by 
\e{\dot{\mm q}_{i,\mathrm{d}}=\mm J_i^\dagger\dot{\mm x}_{i,\mathrm{d}}+\tilde{\mm{J}}_i^\dagger\mm \Lambda \tilde{\mm x}_i}{clik}
%\e{\dot{\mm q}_{i,\mathrm{d}}=\tilde{\mm{J}}_i^\dagger\mm \Lambda \tilde{\mm x}_i}{clik}
with gains $\mm{\Lambda}{=}\mathrm{diag}(\lambda) {\in} \mathbb{R}^{6\times 6}$. Closed-loop inverse kinematics (CLIK) \cite{Chiacchio.1991} avoids the long-term drift whereby we consider only the feedback term, i.e. $\dot{\mm x}_{i,\mathrm{d}}{\equiv}\mm 0$. Including the feedforward component is out of the scope of our work and could further accelerate convergence.
%\e{\Delta{\mm q}_{i,\mathrm{d}}=\dot{\mm q}_{i,\mathrm{d}}\Delta t=\tilde{\mm{J}}_i^\dagger\mm \Lambda \tilde{ \mm x}_i\Delta t.}{clik_final}
\subsection{Shape Fitting using Task-Priority Inverse Kinematics}\label{shapefitting}
\subsubsection{Task-Priority} Forming a desired snake shape requires the positioning of a large number of robot links $i$. In addition to the EE positioning task, the shape-fitting task has to be defined. The task-priority approach \cite{Nakamura.1987,Siciliano.1991} is suitable for fusing these tasks. In contrast to the approach of a weighted sum of the resulting joint angle velocities of each task, a well-defined priority order of the tasks is made possible by null-space projection. For intuitive telemanipulation, the positioning of the robot tip controlled by the user must be executed with high priority. Consequently, shape fitting takes place in the null space of the end-effector task. We use the singularity-robust task-priority framework \cite{Chiaverini.1997} which decouples algorithmic singularities from kinematic singularities by avoiding inversion of the null-space projector. Considering $k$ tasks\footnote{Instead of ${\mm x}_i$ and $\tilde{\mm x}_i$, $\mm{\sigma}_j$ and $\tilde{\mm{\sigma}}_j$ are used as the task variable and task residual respectively, since these do not generally represent the \mbox{$(m_j=6)$-DoF} pose task. Note the difference between task number $j$ and link index $i$ which still denotes the number of the considered segment.} $\mm{\sigma}_j\in\mathbb{R}^{m_j}$ with descending priority order \cite{Baerlocher.1998}, the desired joint angle velocity results in
\e{\dot{\mm q}_{\mathrm{d}}=\tJac{1}^\dagger\mm \Lambda_1 \tilde{ \mm \sigma}_1+\mm N_1\tJac{2}^\dagger\mm \Lambda_2 \tilde{ \mm \sigma}_2+\sum_{j=3}^{k}\mm N_{j-1}^\mathrm{A}\tJac{j}^\dagger\mm \Lambda_j \tilde{ \mm \sigma}_j }{equality-tasks}
with task \mbox{Jacobians $\tJac{j}$}, null space projectors
\e{\mm N_1=\mm I_n-\tJac{1}^\dagger\tJac{1}, \,\mm N_{j-1}^\mathrm{A}=\mm I_n-{\mm J_{j-1}^\mathrm{A}}^\dagger\mm J_{j-1}^\mathrm{A},}{null space}
gains $\mm\Lambda_j{=}\mathrm{diag}(\lambda_j){\in}\mathbb{R}^{m_j\times m_j}$ and augmented Jacobians
\e{\mm J_{j-1}^\mathrm{A}=[\tJac{1}^\mathrm{T},\tJac{2}^\mathrm{T},\hdots,\tJac{j-1}^\mathrm{T}]^\mathrm{T}.}{Jaug}

%%%%%%%VIDEO
%\begin{table}[tbp]
%	\vspace{4mm}
%	\caption{Task definitions: objective, residual, Jacobian, DoF.}
%	\vspace{-4mm}
%	\label{tab:task_def}
%	\begin{center}
%		\begin{tabular}{|c|c|c|c|c|}
%			\hline
%			Task $j$&$\mm{\sigma}_j$&$\tilde{ \mm \sigma}_j$&$\tilde{\mm{J}}_j$&$m_j$\\
%			\hline
%			3T&$\mm{p}_{i}$&$\mm{p}_{i,\mathrm{d}}-\mm{\sigma}_j$&$\mm{J}_{i,\ind{t}}$&3\\
%			\hline
%			1T&$||\mm{p}_{i,\mathrm{d}}-\mm{p}_{i}||_2$&$0-\sigma_j$&$\frac{(\mm{p}_{i,\mathrm{d}}-\mm{p}_{i})^\ind{T}}{-\sigma_j}\mm{J}_{i,\ind{t}}$&1\\
%			\hline
%			3R&$\mm{\varphi}_\mathrm{xyz}(^{0}\mm{R}_i)$&$\mm{\alpha}_\ind{zyx}(^{\ind{d}}\mm{R}_i)$&$\mm{T}_\ind{zyx}\mm{J}_{i,\ind{r}}$&3\\
%			\hline
%			2R&$\mm{\varphi}_\mathrm{xy}(^{0}\mm{R}_i)$&$\mm{\alpha}_\ind{yx}(^{\ind{d}}\mm{R}_i)$&$\mm{T}_\ind{yx}\mm{J}_{i,\ind{r}}$&2\\
%			\hline
%			Fréchet &$d_\mathrm{F}(\mm{P}_\mathrm{d},\mm{P})$&$0-\sigma_j$&$\mm{J}_{\mathrm{F}}$&1\\
%			\hline
%		\end{tabular}
%	\end{center}
%	\vspace{-7mm}
%\end{table}
%%%%%%%VIDEO
\subsubsection{End-Effector Pose} As already mentioned, the tip is positioned on top priority ($j{=}1$) since it is directly controlled by the operator. Additionally, arbitrary safety functions could be defined on higher priorities, e.g. avoiding too abrupt movements due to incorrect input of the operator. Table~\ref{tab:task_def} shows an overview of the task definitions used in this work. The combination of tasks is done by stacking the Jacobians and residuals. For taking a desired EE position $\mm{p}_\mathrm{E,d}\in\mathbb{R}^{3}$, a 3T task is defined which can be extended by an orientation task if necessary. This can also be three-dimensional (3R), so that the full six-DoF pose of the tip can be specified. Alternatively, we consider the relevant case of a pointing task (2R), in which the rotation around the longitudinal axis of the end effector is arbitrary. For image evaluation of the endoscope camera or manipulation with a rotationally symmetrical tool, a specification of the roll angle is obsolete. Moreover, in the considered robot type no rolling motion is possible by the actuators. Omitting the longitudinal rotation as a task (3T2R) thus improves convergence if the given six-DoF pose is not achievable. The Tait-Bryan angle residual of the difference rotation $\tilde{\mm{\varphi}}_\diamond(^{\ind{d}}\mm{R}_\mathrm{E})$ between current orientation $^{\ind{0}}\mm{R}_\mathrm{E}$ and desired orientation $^{\ind{0}}\mm{R}_\ind{d}$ is used as task error \cite{Natale.2003}. Depending on the task, this residual is three-dimensional ($\tilde{\mm{\varphi}}_\mathrm{zyx}$) or two-dimensional ($\tilde{\mm{\varphi}}_\mathrm{yx}$), neglecting the first component. Thus, also two different task Jacobians result by multiplying the transformation $\mm{T}_\ind{zyx}$~(3R) or $\mm{T}_\ind{yx}$~(2R) with the rotational part of the analytical Jacobian $\mm{J}_{i,\ind{r}}$ of link $i$. A detailed derivation can be found in \cite{Schappler.2019}.

\renewcommand{\arraystretch}{1.7}
\begin{table}[btp]
	\vspace{1.5mm}
	\caption{Task definitions: objective, residual, Jacobian, DoF}
	\vspace{-4mm}
	\label{tab:task_def}
	\begin{center}
		\begin{tabular}{|c|c|c|c|c|}
			\hline
			Task $j$&$\mm{\sigma}_j$&$\tilde{ \mm \sigma}_j$&$\tilde{\mm{J}}_j$&$m_j$\\
			\hline
			3T \cite{Antonelli.2014}&$\mm{p}_{i}$&$\mm{p}_{i,\mathrm{d}}-\mm{\sigma}_j$&$\mm{J}_{i,\ind{t}}$&3\\
			\hline
			1T \cite{Antonelli.2014}&$||\mm{p}_{i,\mathrm{d}}-\mm{p}_{i}||_2$&$0-\sigma_j$&$\frac{(\mm{p}_{i,\mathrm{d}}-\mm{p}_{i})^\ind{T}}{-\sigma_j}\mm{J}_{i,\ind{t}}$&1\\
			\hline
			3R \cite{Natale.2003} &$\mm{\varphi}_\mathrm{xyz}(^{0}\mm{R}_i)$&$\tilde{\mm{\varphi}}_\ind{zyx}(^{\ind{d}}\mm{R}_i)$&$\mm{T}_\ind{zyx}\mm{J}_{i,\ind{r}}$&3\\
			\hline
			2R \cite{Schappler.2019} &$\mm{\varphi}_\mathrm{xy}(^{0}\mm{R}_i)$&$\tilde{\mm{\varphi}}_\ind{yx}(^{\ind{d}}\mm{R}_i)$&$\mm{T}_\ind{yx}\mm{J}_{i,\ind{r}}$&2\\
			\hline
			Fréchet &$d_\mathrm{F}(\mm{P}_\mathrm{d},\mm{P})$&$0-\sigma_j$&$\mm{J}_{\mathrm{F}}$&1\\
			\hline
		\end{tabular}
	\end{center}
	\vspace{-7mm}
\end{table}

\subsubsection{Point-to-Point Correspondences} Shape fitting takes place in the null space of the end-effector task. This is possible in two ways, which are presented below. For macroscopic adjustment to a desired configuration it is not necessary to position each link and, moreover, the translational positioning is sufficient. By means of 1T task every $n_\ind{s}$-th link is positioned using the Euclidean distance between current link position $\mm{p}_{i}$ and desired link position $\mm{p}_{i,\mathrm{d}}$. Thereby, $n_\ind{s}$ serves as parameter and will be varied in \ref{shape-fit}. The priority decreases with distance from the EE, so that at link $n{-}n_\ind{s}$ there is the second highest priority ($j{=}2$), at link $n{-}2n_\ind{s}$ the third highest priority ($j{=}3$) and so on. Position $\mm{p}_{n}$ of link $n$ is excluded since the EE task on highest priority is already defined. Nevertheless, it is shown in \ref{shape-fit} that too many tasks negatively influence the convergence. Note that due to the tasks for different links $i$, the Jacobians must be filled with zeros in the last $(n{+}1{-}i)$ columns. Hence, we use the task Jacobian
\e{\tJac{j}=[\tilde{\mm{J}}_j,\mm{0}]\in \mathbb{R}^{m_j \times (n{+}1)} \,\forall j\in\{1,\hdots,k\}}{tJac}
to compute $\dot{\mm{q}}_\ind{d}\in\mathbb{R}^{n{+}1}$ in (\ref{eq:equality-tasks}). This shape-fitting approach is termed ``Point" in Sec. \ref{val}, since point-to-point correspondences between current and desired link positions are used.

\subsubsection{Fréchet Distance as Similarity Measure} Using point-to-point correspondences in both the presented approach and the state-of-the-art \cite{Liljeback.2014, Tang.2020, Wang.2021} is effective as long as the desired positions of all segments are \textit{known}. During FTL locomotion into the target area, the desired shape can be incrementally computed \cite{WilliamsII.1997}. However, for reorientation within the target area by means of pivot movement, the desired positions of all segments are \textit{unknown}: The shape must change due to the kinematic structure in order to establish a new pointing direction $\mm{z}_\ind{d}=\mm{e}_z^{\mathrm{E,d}}$ at constant EE position. When taking the initial shape (before reconfiguration) as the desired shape, point-to-point correspondences are disadvantageous, since the initial shape cannot be maintained if the pointing direction is changed. Especially in the tip area, there is inevitably a large shape deviation which can be seen in Fig.~\ref{fig:cover}(b). To reach the position, the \mbox{feeder $q_1$} must be actuated and the robot has to be moved forward (or backward). Consequently for positioning link $i$, the initial position of this segment is \textit{not reachable}. Instead, the initial positions of following (or preceding) segments should be used as target position for link $i$ to decrease the shape error.

To counter this, we use the Fréchet distance \cite{Frechet.1906} as scalar similarity measure between two curves taking into account the positions and ordering of the points. For fast calculation, the discrete Fréchet distance\footnote{\url{www.github.com/mp4096/discrete-frechet-distance}} $d_\mathrm{F}(\mm{P}_\mathrm{d},\mm{P})$ \cite{Eiter.1994} is used as approximation. The application of this distance measure avoids the need to specify a discrete target position for each segment. Instead, the current shape $\mm{P}$ and the target shape $\mm{P}_\mathrm{d}$ are described as polygonal curves whose vertices are the segment positions $\mm{p}_{i}$ from (\ref{eq:fk}). The initial shape before reconfiguration is taken as target, allowing shape fitting to be realized by minimizing the scalar similarity measure. In contrast to the shape fitting described above using point-to-point correspondences, all initial link positions are thus available to each segment $i$ to reduce the shape error. 

\subsubsection{Implementation Aspects} To minimize the Fréchet distance, only one task has to be defined in the null space of the end-effector task. The computation of the corresponding task Jacobian $\mm{J}_{\mathrm{F}}$ is done by numerical differentiation with angle increment $\delta$ and is presented in Algorithm~\ref{alg:frechet}.
\begin{figure}[t]
	\removelatexerror
	\vspace{2mm}
	\begin{algorithm}[H]
		\caption{Compute Jacobian $\mm{J}_{\mathrm{F}}$ for Fréchet task}\label{alg:frechet}
		{\small 
			\SetKwInOut{Input}{Input}
			\SetKwInOut{Output}{Output}
			\Input{$\mm{q},\delta,\mm{P}_\mathrm{d}$}
			\Output{$\mm{J}_{\mathrm{F}}\in \mathbb{R}^{1 \times (n{+}1)}$}
			$\mm{P}\gets$ Curve from link positions at current configuration $\mm{q}$\;
			$\sigma_j\gets d_\mathrm{F}(\mm{P}_\mathrm{d},\mm{P})$\; 
			
				\For{$i=1$ to $n+1$}{
					$\mm{q}_\delta\gets \mm{q}$\;
					$\mm{q}_\delta\gets $Add small increment $\delta$ to joint angle $i$\; 
					$\mm{P}_\delta\gets$ Curve from link positions at configuration $\mm{q}_{\delta}$\;
					$\sigma_{j,\delta}\gets d_\mathrm{F}(\mm{P}_\mathrm{d},\mm{P}_\delta)$\;
					$\mm{J}_{\mathrm{F},i}\gets (\sigma_{j,\delta}-\sigma_j)/\delta$\;
				}
		}
	\end{algorithm}
	\vspace{-6mm}
\end{figure}
%\begin{figure}[bp]
%	\centerline{\includegraphics[width=0.9\linewidth]{shape_fitting.png}}
%	\caption{Visualization of the presented shape fitting. The initial configuration ($\mm{q}=\mm{0}$) is colored in orange and blue (alternating twist angle $\zeta$). After $n_\ind{iter}=100$ iterations, the robot shape and target shape (green) coincide almost, whereas in the first iterations (gray configurations) most of the shape adjustment takes place.}
%	\label{fig:shape_fitting}
%\end{figure}

%Finally, joint angle limits $q_{i,\ind{min}}\leq q_i\leq q_{i,\ind{max}}$ must be considered as they constrain the workspace. Using the set-based task-priority framework \cite{Moe.2016}, $n{+}1$ set-based tasks would be necessary. Once a joint approaches the limit, the joint becomes inactive and does not exceed the limit. As soon as the algorithm draws the joint away from the limits, the set-based task becomes inactive again. However, $2^{n{+}1}$ modes to be tested maximally in each iteration result for the framework which is infeasible for hyper-redundancy such as $n=30$. Accordingly, the framework is not suitable for online path planning.

Finally, joint angle limits $q_{i,\ind{min}}{\leq} q_i{\leq} q_{i,\ind{max}}$ are considered. When a violation of joint $i{=}\kappa$ occurs, the corresponding joint angle $q_\kappa$ is pragmatically set to its limit. This scales one entry in $\dot{\mm q}_{\mathrm{d}}$ resulting in no more null-space motion \cite{Baerlocher.2004}. Recalculating the CLIK step (\ref{eq:equality-tasks}) is required, where column $\kappa$ of all task Jacobians $\tJac{j}$ is removed \cite{Leibrandt.2017}.  The desired position change per iteration scaled by time step $\Delta t$ results in \mbox{$\Delta{\mm q}_{\mathrm{d},\mm{\kappa}}=\dot{\mm q}_{\mathrm{d},\mm{\kappa}}\Delta t$} and includes only the motion of the active joints. The vector $\mm{\kappa}$ contains binary entries and serves as an overview of which joints are active~(${=}1$) and which are inactive~(${=}0$). Within a CLIK iteration, this procedure is repeated until no limits are exceeded. After acceptance of an iteration, all joints are set back to the initial state $\mm{\kappa}{=}\mm{\kappa}_\ind{init}$. In the case of only active joints, $\mm{\kappa}_\ind{init}=[1,\hdots,1]^\ind{T}\in\mathbb{R}^{n{+}1}$ thus holds for $n$ rotational actuators and one feeder. 

Algorithm~\ref{alg:shapefitting} introduces our joint-limit-aware implementation of the shape fitting as pseudo-code. The desired end-effector pose can be specified via $^{0}\mm{T}_\ind{E,d}$. Depending on the shape-fitting approach, the Jacobians $\tJac{1..k}$ and residuals $\tilde{\mm \sigma}_{1..k}$ vary. Based on this, the unified strategy for snake telemanipulation is presented in the following section.

\begin{figure}[t]
	\removelatexerror
	\vspace{2mm}
	\begin{algorithm}[H]
		\caption{Shape fitting using task-priority}\label{alg:shapefitting}
		{\small 
			\SetKwInOut{Input}{Input}
			\SetKwInOut{Output}{Output}
			\Input{$\mm{q},\mm{P}_{\mathrm{d}},^{0}\mm{T}_\ind{E,d},\mm{\kappa}_\mathrm{init}$}
			\Output{${\mm q}_{\mathrm{d}}$}
			\For{$\mathrm{iter}=1$ to $n_\mathrm{iter}$}{
				$\tJac{1..k}, \tilde{\mm \sigma}_{1..k}\gets$ Update with $\mm{q}$\;
				$\mathrm{flag}$ $\gets$ Set joint-limit flag to 0\;
				$\mm{\kappa}$ $\gets \mm{\kappa}_\mathrm{init}$  Reset joint states\;
				\While{$\mathrm{flag}=0$}{
					$\dot{\mm q}_{\mathrm{d},\mm{\kappa}}\gets$ CLIK step (\ref{eq:equality-tasks}) with active joints $\mm{\kappa}$\;
					$\Delta{\mm q}_{\mathrm{d},\mm{\kappa}}\gets \dot{\mm q}_{\mathrm{d},\mm{\kappa}}\Delta t$ with limited step size\;
					$\kappa\gets$ First joint index with exceeded limit or 0\;
					\uIf{$\kappa=0$}{
						$\mathrm{flag}\gets$ Set joint-limit flag to 1\;
						Add $\Delta\mm{q}_{\mathrm{d},\mm{\kappa}}$ to active joints $\mm{\kappa}$ in $\mm{q}$\;}
					\Else {Lock $q_\kappa$ on limit in $\mm{q}$\;
						$\mm{\kappa}$ $\gets$ Set joint $\kappa$ to 0\;
					}
				}
			}
				${\mm q}_{\mathrm{d}}\gets\mm{q}$\;
			
		}
	\end{algorithm}
	\vspace{-6mm}
\end{figure}
\begin{figure}[t]
	\removelatexerror
	\vspace{2mm}
	\begin{algorithm}[H]
		\caption{SnakeTTP}\label{alg:snakettp}
		{\small 
			\SetKwInOut{Input}{Input}
			\SetKwInOut{Output}{Output}
			\Input{$\mm{q},b_1,b_2,\m{R}{}{B}{}{S}$}
			\Output{${\mm q}_{\mathrm{d}}$}
			$\mm{\varphi}_\mathrm{xy}\gets$ Pitch and yaw angles of $\m{R}{}{B}{}{S}$\;
			$\mm{\kappa}_\mathrm{init}\gets$ Set states of active joints to 1\;
			\uIf{$b_1$}
				{\tcp{FTL locomotion}
				$\mm{q}\gets$ Update pointing direction using $\mm{\varphi}_\mathrm{xy}$\;
				$\mm{P}_{\mathrm{d}}\gets$ Curve from desired link positions \cite{WilliamsII.1997}\;
				$^{0}\mm{T}_\ind{E,d}\gets$ Desired EE pose with $\mm{\varphi}_\mathrm{xy}$ and $\mm{P}_{\mathrm{d}}$\;
				${\mm q}_{\mathrm{d}}\gets$ Shape fitting (point-to-point correspondences)\;
			}
			\uElseIf{$b_2$}
				{\tcp{Pivot reorientation}
				$\mm{P}_{\mathrm{d}}\gets$ Curve from current link positions with $\mm q$\;
				$^{0}\mm{T}_\ind{E,d}\gets$ Desired EE pose with $\mm{\varphi}_\mathrm{xy}$ and $\mm{P}_{\mathrm{d}}$\;
				${\mm q}_{\mathrm{d}}\gets$ Shape fitting (Fréchet distance)\;
			}
			\Else{
				\tcp{Telemanipulate last module}
				$\mm{q}\gets$ Update pointing direction using $\mm{\varphi}_\mathrm{xy}$\;
				${\mm q}_{\mathrm{d}}\gets\mm{q}$\;
			}
		}
	\end{algorithm}
	\vspace{-6mm}
\end{figure}
\subsection{Telemanipulated Locomotion and Reorientation}\label{ftlform}
%The approach for real-time knot-tying simulation \cite{Brown.2004} used in FABRIK \cite{Aristidou.2011} would be possible to determine those quantities. However, this results in movements of the base in any direction, which is not feasible when using a linear, one-dimensional feeder.
One contribution of our work is a unified strategy for telemanipulated locomotion and reorientation using task-priority inverse kinematics. Algorithm \ref{alg:snakettp} presents the approach with user input $\{b_1,b_2,\m{R}{}{B}{}{S}\}$ and will be denoted SnakeTTP (snake telemanipulation using task-priority) in the following. Intuitive telemanipulation is realized by the six-DoF input device shown in Fig.~\ref{fig:cover}(a). It would be possible to control both feeder $q_1$ and pointing direction $\mm{e}_z^{\mathrm{E}}$ by changing the pose of the stylus $\m{T}{}{B}{}{S}$. However, for the one-DoF feeder it is convenient to control it separately via buttons \mbox{$b_\diamond\in\{0,1\}$}. This allows the user to focus completely on setting the desired pointing direction by changing orientation $\m{R}{}{B}{}{S}$. Thus, a change of $\mm{e}_z^{\ind{S}}$ by the user directly results in a changed pointing direction of the snake robot $\mm{e}_z^{\mathrm{E}}$ in steady state. Depending on the buttons pressed by the operator, there are three different cases which are presented below.

If no button is pressed, no locomotion takes place. Only the last module is controlled, resulting in a reorientation by changing the pointing direction. Determining the pitch and yaw angles $\mm{\varphi}_\mathrm{xy}(\m{R}{}{B}{}{S})$, the joint angles of the last module $\{q_n,q_{n+1}\}$ can be changed under consideration of joint angle limits. Consequently, no pivot movement occurs in this type of reorientation and the EE position changes.

Telemanipulated FTL motion is done by pressing \mbox{button $b_1$}. An incremental calculation of each target position $\mm{p}_{i,\ind{d}}$ for the next time step based on the current configuration is necessary. We use a successive approach starting at the tip \cite{WilliamsII.1997}. The points $\mm{p}_{\mathrm{E,d}}$ and $\mm{p}_{n,\ind{d}}$ describe the desired position of the end effector and the link $n$, respectively. Estimating those quantities is realized by moving $\Delta s$ in the pointing direction $\mm{e}_z^{\mathrm{E}}$. Since two rotational actuators are necessary for spatial movement, a target position is consecutively determined for every second link. The position~$\mm{p}_{i}$ is calculated by placing the center of a sphere of radius $r_\ind{s}$ at $\mm{p}_{i+2}$ and computing the intersection between the sphere and the robot structure represented as a piecewise linear curve. Thereby, $r_\ind{s}$ is the distance between the link $i$ and link $i{+}2$. The position $\mm{p}_{i}$ is then used as the new sphere center to determine $\mm{p}_{i-2}$. The calculation continues downward along the robot structure until the base is reached. The result is a matrix 
\e{\mm{P}_\ind{d,all}=[\mm{p}_{\mathrm{E,d}},\mm{p}_{n,\ind{d}},\mm{p}_{n-2,\ind{d}},\hdots,\mm{p}_{2,\ind{d}}]^\ind{T}}{P_d}
containing target positions for all $n/2$ modules and one EE position. The number of positions used for shape fitting is controlled via parameter $n_\ind{s}$ to obtain $\mm{P}_\ind{d}$ (see \ref{shapefitting}). 

After the robot moved into the target area, the operator can control the tip by means of pivot movement (button $b_2$). The target shape results from the current link positions to reduce the shape change due to the new pointing direction. Shape fitting is performed via 3T2R EE-task and minimization of the Fréchet distance by null-space projection.

There is one final remark to be made on the implementation. In our work we consider a snake robot which moves out of a rigid tube. This is indicated in Fig.~\ref{fig:cover}(a) in the bottom part of the snake robot by the gray cylinder. Accordingly, the number of available joints increases with advanced feed position $q_1$. We implement this analogously to the consideration of the joint angle limits by deleting columns of the task Jacobians. At the beginning, two rotational actuators are available for a spatial motion, so that $\mm{\kappa}_\ind{init}=[1,\mm{0},1,1]^\ind{T}$ holds. Once $q_1$ has increased by one actuator height, the third actuator is activated and $\mm{\kappa}_\ind{init}=[1,\mm{0},1,1,1]^\ind{T}$ applies. Thus in the vector $\mm{\kappa}_\ind{init}$ with binary entries, the available actuators are considered in each time step. 

%Second, it takes some time for the controllers to reach the desired robot configuration $\mm{q}_\ind{d}$. During this time, no recomputation of the inverse kinematics should take place. Therefore, the old solution ${\mm{q}}_\ind{d,old}$ is updated with the new pointing direction $\mm{\phi}_\mathrm{xy}$ and passed to the controllers again as the target configuration. It allows the operator to update the pointing direction at each time step. This is repeated until approximately the desired feed position has been taken and $q_\mathrm{1}\approx{q}_{\mathrm{d,old,1}}$ applies.
\section{Validation}\label{val}
We validate our method in three experiments using a virtual snake robot with $n{=}30$ rotational actuators, symmetric joint angle limits $q_{i,\ind{max}}{=}{-}q_{i,\ind{min}}{=}\ang{30}\,\forall i{>}1$ and an actuator height \mbox{$h{=}\SI{10}{\milli\meter}$}. For the EE positioning errors $\{\mathcal{X}_\mathrm{3T}, \mathcal{X}_\mathrm{2R}, \mathcal{X}_\mathrm{3R}\}$, the norm is calculated from the respective task error $||\tilde{\mm \sigma}_{\mathrm{E}}||_2$ from Table~\ref{tab:task_def}. Position and shape errors are normalized to the actuator height $h$ to make these values more comprehensible. The proposed method was implemented using Matlab/Simulink R2021b on a 2.5\,GHz Intel Core i5-10300H CPU with 16\,GB of RAM running Windows. The functions for computing the forward kinematics and Jacobians were mex-compiled. Since we do not consider an instrument at the distal end, the translational part of the tool transformation $^{n{+}1}\mm{T}_\ind{E}$ disappears. Code\footnote{\url{www.github.com/tlhabich/snakettp}} to reproduce the results is publicly available. 

%First, the presented shape fitting is compared to a state-of-the-art method. After proving the functionality, a user study validates the proposed telemanipulation strategy. For error estimation the Euclidean distance $\mathcal{X}_i$ from each link position $\mm{r}_i$ to the desired shape (perpendicular to the longitudinal robot axis) is determined. As metric of shape fitting, the mean shape error
%\e{\mathcal{X}_\mathrm{avg}=\frac{1}{n+1}\sum_{i=1}^{n+1}\mathcal{X}_i}{xavg}
%and the maximum distance
%\e{\mathcal{X}_\mathrm{max}=\max_{1\leq i\leq {n+1}} \mathcal{X}_i}{xavg}
%are considered. 
\subsection{Shape Fitting}\label{shape-fit}
\begin{figure}[bp]
	\vspace{1.5mm}
	\centerline{\includegraphics[width=\linewidth]{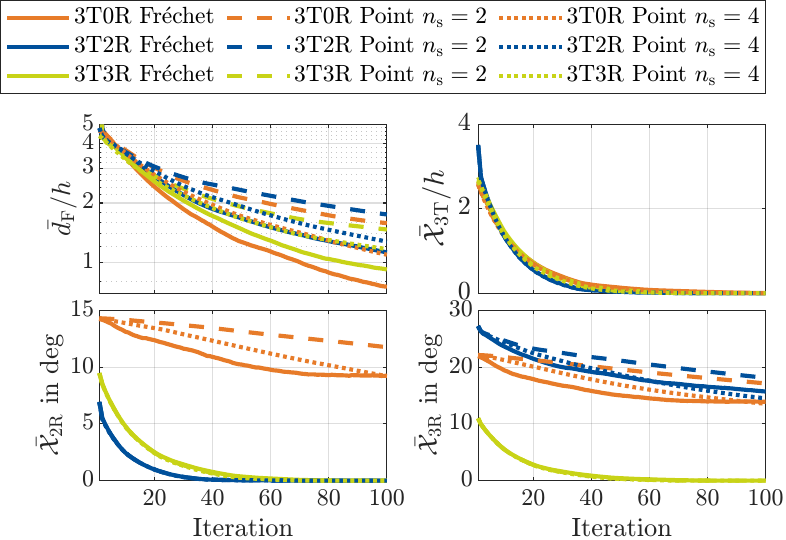}}
	\caption{Convergence of shape and pose errors averaged for 100 random desired shapes. One iteration takes on average $\SI{1.5}{\milli\second}$.}
	\label{fig:convergence_frechet}
	\vspace{-1.5mm}
\end{figure}
\begin{figure*}[tbp]
	\vspace{1.5mm}
	\centerline{\includegraphics[width=\linewidth]{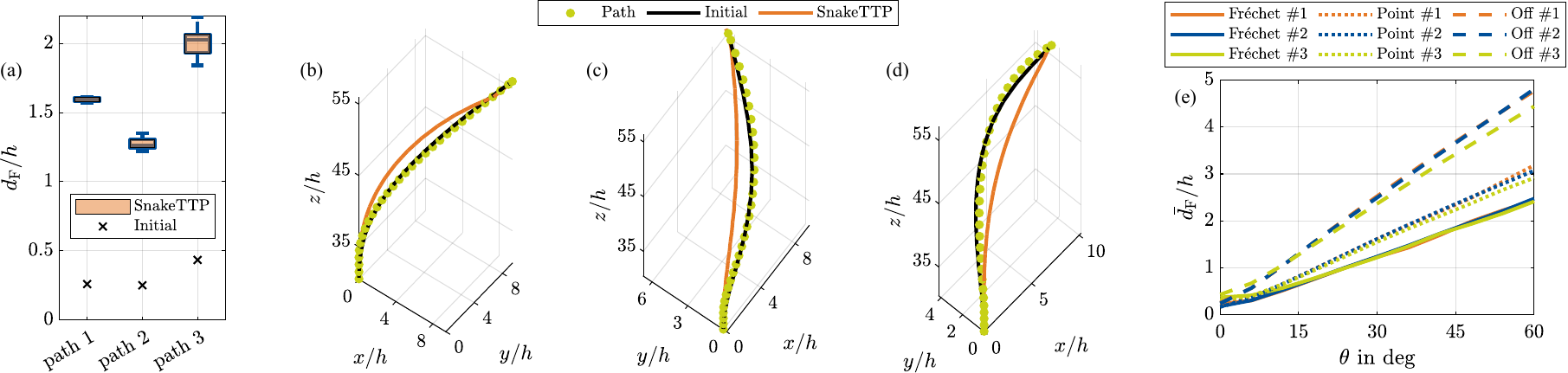}}
	\caption{(a) Normalized shape error at the end of the telemanipulation run using SnakeTTP compared to the shape-fitting method with knowledge of the entire path (Initial). (b)--(d) Spatial visualization of the three paths, the final configurations of an exemplary study participant (SnakeTTP) and the shape fitting results given knowledge of the entire paths (Initial). (e) Results of the reorientation experiment for shape fitting using Fréchet distance, point-to-point correspondences and with disabled shape fitting for three paths. The opening angle of the viewing cone $\theta$ was increased equidistantly.}
	\label{fig:results_tele_pivot}
	\vspace{-1.5mm}
\end{figure*}
In this section the accuracy and convergence of the two formulated task-priority approaches are compared. A set of 100 random target configurations $\mm{q}_\ind{d}$ without feeder ($q_1\equiv0$) is generated and desired link positions $\mm{P}_\mathrm{d}$ are computed using (\ref{eq:fk}). Each method gets these points as input and starts in an identical configuration ($\mm{q}=\mm{0}$). Tuning the \mbox{gains $\lambda_j$} is out of the scope of this paper. Thus, $\lambda_j=1\,\forall j$ applies.

Figure~\ref{fig:convergence_frechet} plots the shape and pose errors averaged ($\bar{\diamond}$) for 100 shape-fitting tasks with increasing number of iterations. All strategies enable fast convergence to a desired pose. Depending on the definition of the end-effector task, the respective task error disappears. Moreover, the shape fitting done in the null space achieves for the worst method a mean Fréchet distance smaller than two actuator heights. The proposed shape fitting using the Fréchet distance shows fastest convergence and realizes a mean shape error of less than $0.8h$. Since only one scalar task is performed in the null space of the EE task, this fast convergence is possible. For point-to-point correspondences, a variation of $n_\ind{s}$ shows that the fitting is improved if every fourth link ($n_\ind{s}{=}4$) is positioned. For $n_\ind{s}{=}2$, the task dimension is significantly larger which affects the convergence. Hence, we will use $n_\ind{s}{=}4$ in the following. The 3T task will be defined for the EE during FTL motion which shows fast shape convergence due to the low task dimension.
\subsection{Telemanipulation Study}\label{tele-perf}
The second experiment validates the telemanipulated FTL movement. A user study was conducted with 14 participants, all of them male engineers without any endoscopy experience. MuJoCo \cite{Todorov.2012} was used as a simulation environment for the user to receive visual feedback through an endoscope camera as shown in Fig.~\ref{fig:cover}(a). Three different target paths were given by green spheres for the subject to follow. The increment in pointing direction is set to $\Delta {s}=\frac{h}{20}$ and, therefore, few iterations are required for convergence from the current configuration to the target shape. Thus, the number of maximum iterations can be reduced to $n_\mathrm{iter} {=} 50$. The step size is limited by $\Delta t{=}\SI{1}{\second}$. To increase robustness, for the maximum change of the rotational actuators $\Delta q_\mathrm{r,max}{=} \frac{15\degree}{n_\mathrm{iter}}$ and for the feeder $\Delta q_\mathrm{1,max}{=} \frac{2\Delta s}{n_\mathrm{iter}}$ applies per CLIK step. We use 3D Systems' Geomagic Touch as input device and the PhanTorque library \cite{Aldana.2014} for Simulink integration.

The results of the study are illustrated in Fig.~\ref{fig:results_tele_pivot}(a)--(d). For all three paths, the normalized Fréchet distance in the final configuration was determined. In addition, exemplary from one randomly chosen participant the final shapes are visualized spatially. As a comparison, the result of our shape fitting is also visualized, which receives the \textit{entire target shape} as input (denoted as ``Initial" for the next section). As expected, with knowledge of the whole path, a significantly better shape fitting is possible since the shape fitting is not created incrementally via user input. Depending on the curvature characteristics, the shape-fitting and telemanipulation performance differs. Nevertheless, macroscopic path-tracking was achieved in all telemanipulation tasks.

\subsection{Reorientation within Target Area}\label{reorientation}
In a final experiment the core of the paper will be validated: a pivot motion in order to reorient within the target area \textit{keeping the shape change as small as possible}. For this purpose, the three target paths of the study are used again. The shape-fitting result denoted as ``Initial" from Fig.~\ref{fig:results_tele_pivot}(b)--(d) with the knowledge of the entire desired shape is given as the \textit{initial} shape. Thus, there is a very small shape error at the beginning. The pointing direction of the initial configuration $\mm{z}_0{=}\mm{e}_z^{\mathrm{E}}$ is transformed into a new desired pointing direction $\mm{z}_\ind{d}{=}\mm{e}_z^{\mathrm{E,d}}$ using spherical coordinates $\{\theta,\phi\}$ to describe a cone-shaped area for reorientation. Thereby, $0{\leq}\theta{\leq}\theta_\ind{max}$ represents the opening angle of the cone and $0{\leq}\phi{\leq} 2\pi$ the azimuth angle. Starting at $\theta{=}0\degree$, the opening angle is increased in ten equidistant steps to $\theta_\ind{max}{=}60\degree$. At each opening angle step, a complete rotation is specified in the same way by means of $\phi$. Thus, $11^2$ desired pointing directions $\mm{z}_\ind{d}$ are specified for each path. By means of a 3T2R task for the EE, this pointing direction and the constant position can be defined to realize a pivot movement\footnote{Illustration of the procedure: \url{https://youtu.be/oIpbxpJPoyQ}}. 

Figure~\ref{fig:results_tele_pivot}(e) shows the results of the experiment. The larger the deviation between the desired and initial pointing direction, the larger the Fréchet distance. The curve is almost identical for all three paths. However, using the Fréchet task allows for less shape change than using point-to-point correspondences. For the maximum orientation change considered, the shape deviation using the Fréchet approach is on average $20.1\%$ smaller compared to the reference. Figure~\ref{fig:cover}(b) shows such a reconfiguration for path $\#1$.

%\begin{figure}[tbp]
%	\vspace{1.5mm}
%	\centerline{\includegraphics[width=\linewidth]{results_pivot_left_theta_max_60deg_n_theta_10_n_phi_10right_theta60deg_phi288deg.pdf}}
%	\caption{bla.}
%	\vspace{-1.5mm}
%	\label{fig:pivot_path3}
%\end{figure}
\section{Conclusions}\label{conclusions}
This paper presents SnakeTTP, a unified algorithm for intuitive telemanipulation realizing locomotion and pivot reorientation for endoscopic tasks. The new method based on task-priority inverse kinematics allows different position and orientation specifications at highest priority and shape fitting within the null space. The first experiment validates its fast convergence to desired EE poses and snake shapes. The proposed shape fitting using the Fréchet distance shows fastest convergence and realizes a mean shape error less than one actuator height. In a user study with 14 participants, it is shown that SnakeTTP enables online-planned locomotion with macroscopic path tracking. The last experiment demonstrates the benefit of our approach during reorientation within the target area. The novel shape-fitting approach based on the Fréchet distance reduces the shape error up to $20.1\%$ in contrast to the classical strategy using Euclidean distance between current and desired link positions.
\addtolength{\textheight}{-4.5cm}   % This command serves to balance the column lengths

%%%%%%%%%%%%%%%%%%%%%%%%%%%%%%%%%%%%%%%%%%%%%%%%%%%%%%%%%%%%%%%%%%%%%%%%%%%%%%%%

%%%%%%%%%%%%%%%%%%%%%%%%%%%%%%%%%%%%%%%%%%%%%%%%%%%%%%%%%%%%%%%%%%%%%%%%%%%%%%%%

%%%%%%%%%%%%%%%%%%%%%%%%%%%%%%%%%%%%%%%%%%%%%%%%%%%%%%%%%%%%%%%%%%%%%%%%%%%%%%%%
\section*{Acknowledgment}
The authors acknowledge the support of this project by the German Research Foundation (Deutsche Forschungsgemeinschaft) under grant number 433586601.
%%%%%%%%%%%%%%%%%%%%%%%%%%%%%%%%%%%%%%%%%%%%%%%%%%%%%%%%%%%%%%%%%%%%%%%%%%%%%%%%
\bibliographystyle{IEEEtran}
\bibliography{literatur}

\end{document}